\documentclass[12pt]{article}
\usepackage{longtable}
\usepackage{makecell}
\usepackage{hyperref}
\usepackage{graphicx}
\usepackage{amsmath}
\usepackage{booktabs}
\usepackage{amsfonts}
\usepackage{tabulary}
\usepackage{ragged2e}
\usepackage{floatrow}
\usepackage[letterpaper,top=1in,bottom=1in,left=1in,right=1in,marginparwidth=1.75cm]{geometry}
\usepackage[utf8]{inputenc}
\usepackage[round]{natbib}
\usepackage{float}
\usepackage{booktabs} 
\usepackage{xcolor}
\definecolor{bg}{rgb}{0.95,0.95,0.95}
\usepackage[frozencache=true]{minted}
\usepackage{tabularx}  
\newcolumntype{Y}{>{\centering\arraybackslash}X}

\begin{document}

\title{PyCFRL: A Python library for counterfactually fair offline reinforcement learning via sequential data preprocessing\footnote{\texttt{PyCFRL} Github repository: \url{https://github.com/JianhanZhang/PyCFRL}; \texttt{PyCFRL} documentation: \url{https://pycfrl-documentation.netlify.app/}.}}




\author{Jianhan Zhang\textsuperscript{1}, 
Jitao Wang\textsuperscript{2}, 
Chengchun Shi\textsuperscript{3}, John D. Piette\textsuperscript{4}, \\
Donglin Zeng\textsuperscript{2}, 
Zhenke Wu\textsuperscript{2,}\footnote{Corresponding author: Zhenke Wu (zhenkewu@umich.edu).} \\
\textsuperscript{1}Department of Statistics, University of Michigan, USA\\
\textsuperscript{2}Department of Biostatistics, University of Michigan, USA \\
\textsuperscript{3}Department of Statistics, London School of Economics, UK \\
\textsuperscript{4}Department of Health Behavior and Health Equity, School of Public Health, \\
University of Michigan, USA
}
\maketitle

\section{Summary}
Reinforcement learning (RL) aims to learn and evaluate a sequential
decision rule, often referred to as a “policy”, that maximizes
expected discounted cumulative rewards to optimize the population-level benefit in an environment across possibly infinitely many time steps. RL has gained popularity in fields such as healthcare, banking, autonomous driving, and, more recently, large language model fine-tuning. However, the sequential decisions made by an RL algorithm, while optimized to maximize overall population benefits, may disadvantage certain individuals who are in minority or socioeconomically disadvantaged groups. A fairness-unaware RL algorithm learns an optimal policy that makes decisions based on the \textit{observed} state variables. However, if certain values of the sensitive attribute influence the state variables and lead the policy to systematically withhold certain actions from an individual, unfairness will result. For example, Hispanics may under-report their pain levels due to cultural factors, misleading a fairness-unaware RL agent to assign less therapist time to these individuals \citep{piette2023powerED}. Deployment of RL algorithms without careful fairness considerations can raise concerns and erode public trust in high-stakes settings.

To formally define and address the fairness problem in the novel sequential decision-making settings, \cite{wang2025cfrl} extended the concept of single-stage counterfactual fairness (CF) in a structural causal framework \citep{kusner2018cf} to the
multi-stage setting and proposed a data preprocessing algorithm that
ensures CF. A policy is counterfactually fair if, at every time step, the probability of assigning any action does not change had the individual's sensitive attribute taken a different value, while holding constant other historical exogenous variables and actions. In this light, the data preprocessing algorithm ensures CF by constructing new state variables that are not impacted by the sensitive attribute(s). Reward preprocessing is also conducted, but with a different purpose to improve the value of the learned optimal policy rather than to ensure CF. We refer interested readers to \cite{wang2025cfrl} for more technical details.

The \texttt{PyCFRL} library implements the data preprocessing algorithm proposed by \cite{wang2025cfrl} and provides functionalities to evaluate the value (expected discounted cumulative reward) and counterfactual unfairness level achieved by 
any given policy. Here, "CFRL" stands for "Counterfactual Fairness in Reinforcement Learning". The library produces preprocessed trajectories that can be used by
an off-the-shelf offline RL algorithm, such as fitted Q-iteration (FQI) \citep{riedmiller2005fqi}, to learn an optimal CF
policy. The library can also simply read in any policy following a required format and return its
value and counterfactual unfairness level in the environment of interest, where the environment can be either pre-specified or learned from the data.

\section{Statement of Need}
Many existing Python libraries implement algorithms designed to ensure fairness in machine learning. For example, \texttt{Fairlearn} \citep{weerts2023fairlearn} and \texttt{aif360} \citep{aif360-oct-2018} provide tools for mitigating bias in single-stage machine learning predictions under statistical association-based fairness criteria such as demographic parity and equal opportunity. However, existing libraries do not focus on counterfactual fairness, which defines an individual-level fairness concept from a causal perspective, and they cannot be easily extended to the general RL setting. Scripts available from \texttt{ml-fairness-gym} \citep{fairness_gym} allow users to simulate unfairness in sequential decision-making, but they neither implement algorithms that reduce unfairness nor address CF. To our knowledge, \cite{wang2025cfrl} is the first work to study CF in RL. Correspondingly, \texttt{PyCFRL} is also the first code library to address CF in the RL setting.

The contribution of \texttt{PyCFRL} is two-fold. First, \texttt{PyCFRL} implements a data preprocessing algorithm that ensures CF in offline RL. For each individual in the data, the preprocessing algorithm sequentially estimates and concatenates the counterfactual states under different sensitive attribute values with the observed state at each time point into a new state vector. The preprocessed data can then be directly used by existing RL algorithms for policy learning, and the learned policy will be counterfactually fair up to finite-sample estimation accuracy. Second, \texttt{PyCFRL} provides a platform for assessing RL policies based on CF. After passing in any policy and a data trajectory from the environment of interest, users can estimate the value and counterfactual unfairness level achieved by the policy in the environment of interest. 

\section{High-level Design}
The \texttt{PyCFRL} library is composed of 5 major modules. The functionalities of the modules are summarized in the table below.

\begin{table}[h!]
\centering
\begin{tabular}{l|p{10cm}}
\toprule
\textbf{Module} & \textbf{Functionalities} \\
\midrule
\texttt{reader} & Implements functions that read tabular trajectory data from either a \texttt{.csv} file or a \texttt{pandas.Dataframe} into an array format required by \texttt{PyCFRL}. Also implements functions that export trajectory data to either a \texttt{.csv} file or a \texttt{pandas.Dataframe}. \\
\texttt{preprocessor} & Implements the data preprocessing algorithm introduced in \cite{wang2025cfrl}. \\
\texttt{agents} & Implements an FQI algorithm \citep{riedmiller2005fqi}, which learns RL policies and makes decisions based on the learned policy. Users can also pass a preprocessor to the FQI; in this case, the FQI will be able to take in unpreprocessed trajectories, internally preprocess the input trajectories, and directly output counterfactually fair policies. \\
\texttt{environment} & Implements a synthetic environment that produces synthetic data as well as a simulated environment that estimates and simulates the transition dynamics of the unknown environment underlying some real-world RL trajectory data. Also implements functions for sampling trajectories from the synthetic and simulated environments. \\
\texttt{evaluation} & Implements functions that evaluate the value and counterfactual unfairness level of a policy. Depending on the user's needs, the evaluation can be done either in a synthetic environment or in a simulated environment. \\
\bottomrule
\end{tabular}
\caption{Modules in the \texttt{PyCFRL} library and their functionalities.}
\label{tab:modules}
\end{table}

A general \texttt{PyCFRL} workflow is as follows: First, simulate trajectories using \texttt{environment} or read in trajectories using \texttt{reader}. Then, train a preprocessor using \texttt{preprocessor} and preprocess the training trajectory data. After that, pass the preprocessed trajectories into the FQI algorithm in \texttt{agents} to learn a counterfactually fair policy. Finally, use functions in \texttt{evaluation} to evaluate the value and counterfactual unfairness level of the trained policy. 

In addition, \texttt{PyCFRL} also provides tools to check for potential non-convergence that may arise during the training of neural networks, FQI, or fitted-Q evaluation (FQE). More discussions about the sources, checks, and fixes of non-convergence in \texttt{PyCFRL} can be found in the \href{https://pycfrl-documentation.netlify.app/tutorials/common_issues}{"Common Issues"} section of the documentation.

\section{Data Example}
We provide a data example showing how \texttt{PyCFRL} learns counterfactually fair policies from real-world trajectory data with unknown underlying transition dynamics. The example demonstrates policy learning and evaluation of both value and unfairness levels. This represents just one of many possible workflows. \texttt{PyCFRL} can also generate synthetic trajectory data and evaluate custom preprocessing methods. See the \href{https://pycfrl-documentation.netlify.app/tutorials/example_workflows}{"Example Workflows"} documentation for more examples.

We record the computing times of different workflows under different combinations of the number of individuals ($N$) and the length of horizons ($T$) in the \href{https://pycfrl-documentation.netlify.app/introduction/computing_times}{"Computing Times"} section of the \texttt{PyCFRL} documentation. For example, under $N=500$ and $T=10$, the workflow presented in this data example ran for 378.6 seconds on average in our computing environment.

\subsection{Load Data}

In this demonstration, we use an offline trajectory generated from a \texttt{SyntheticEnvironment} following some pre-specified transition rules. Although the data is actually synthesized, we treat it as if it is from some unknown environment for pedagogical convenience.

The trajectory contains 500 individuals (i.e., $N=500$) and 10 transitions (i.e., $T=10$). The sensitive attribute variable and the state variable are both univariate. The sensitive attribute is binary ($0$ or $1$). The actions are also binary ($0$ or $1$) and are sampled using a behavior policy that selects $0$ or $1$ randomly with equal probability. The trajectory is stored in a tabular format in a \texttt{.csv} file. We use \texttt{read\_trajectory\_from\_csv()} to load the trajectory from the \texttt{.csv} format into the array format required by \texttt{PyCFRL}.

\begin{minted}[fontsize=\small, bgcolor=bg]{python}
zs, states, actions, rewards, ids = read_trajectory_from_csv(
    path='../data/sample_data_large_uni.csv', z_labels=['z1'], 
    state_labels=['state1'], action_label='action', reward_label='reward', 
    id_label='ID', T=10)
\end{minted}

We then split the trajectory data into a training set (80\%) and a testing set (20\%) using scikit-learn's \texttt{train\_test\_split()}. The training set is used to train the counterfactually fair policy, while the testing set is used to evaluate the value and counterfactual unfairness level achieved by the policy.

\begin{minted}[fontsize=\small, bgcolor=bg]{python}
(zs_train, zs_test, states_train, states_test, 
 actions_train, actions_test, rewards_train, rewards_test
) = train_test_split(zs, states, actions, rewards, test_size=0.2)
\end{minted}

\subsection{Train Preprocessor \& Preprocess Trajectories}

We now train a \texttt{SequentialPreprocessor} and preprocess the trajectory. The \texttt{SequentialPreprocessor} ensures the learned policy is counterfactually fair by constructing new state variables that are not impacted by the sensitive attribute. Due to limited trajectory data, the data to be preprocessed will also be the data used to train the preprocessor, so we set \texttt{cross\_folds=5} to reduce overfitting. In this case, \texttt{train\_preprocessor()} will internally divide the training data into 5 folds, and each fold is preprocessed using a model that is trained on the other 4 folds. We initialize the \texttt{SequentialPreprocessor}, and \texttt{train\_preprocessor()} will take care of both preprocessor training and trajectory preprocessing.

\begin{minted}[fontsize=\small, bgcolor=bg]{python}
sp = SequentialPreprocessor(z_space=[[0], [1]], num_actions=2, cross_folds=5, 
                            mode='single', reg_model='nn')
states_tilde, rewards_tilde = sp.train_preprocessor(
    zs=zs_train, xs=states_train, actions=actions_train, rewards=rewards_train)
\end{minted}

\subsection{Counterfactually Fair Policy Learning}

Next, we train a counterfactually fair policy using the preprocessed data and \texttt{FQI} with \texttt{sp} as its internal preprocessor. By default, the input data will first be preprocessed by \texttt{sp} before being used for policy learning. However, since the training data \texttt{states\_tilde} and \texttt{rewards\_tilde} are already preprocessed in our case, we set \texttt{preprocess=False} during training so that the input trajectory will not be preprocessed again by the internal preprocessor (i.e., \texttt{sp}).

\begin{minted}[fontsize=\small, bgcolor=bg]{python}
agent = FQI(num_actions=2, model_type='nn', preprocessor=sp)
agent.train(zs=zs_train, xs=states_tilde, actions=actions_train, 
            rewards=rewards_tilde, max_iter=100, preprocess=False)
\end{minted}

\subsection{\texttt{SimulatedEnvironment} Training}

Before moving on to the evaluation stage, there is one more step: We need to train a \texttt{SimulatedEnvironment} that mimics the transition rules of the true environment that generated the training trajectory, which will be used by the evaluation functions via Monte Carlo. To do so, we initialize a \texttt{SimulatedEnvironment} and train it on the whole trajectory data (i.e., training set and testing set combined).

\begin{minted}[fontsize=\small, bgcolor=bg]{python}
env = SimulatedEnvironment(num_actions=2, state_model_type='nn', 
                           reward_model_type='nn')
env.fit(zs=zs, states=states, actions=actions, rewards=rewards)
\end{minted}

\subsection{Value and Counterfactual Unfairness Level Evaluation}

We now use \texttt{evaluate\_value\_through\_fqe()} and \texttt{evaluate\_fairness\_through\_model()} to estimate the value and counterfactual unfairness level achieved by the trained policy when interacting with the environment of interest, respectively. The counterfactual unfairness level is represented by a metric from 0 to 1, with 0 representing perfect fairness and 1 indicating complete unfairness. We use the testing set for evaluation.

\begin{minted}[fontsize=\small, bgcolor=bg]{python}
value = evaluate_reward_through_fqe(zs=zs_test, states=states_test, 
    actions=actions_test, rewards=rewards_test, policy=agent, model_type='nn')
cf_metric = evaluate_fairness_through_model(env=env, zs=zs_test, 
                                            states=states_test, 
                                            actions=actions_test, policy=agent)
\end{minted}

The estimated value is $7.358$ and the CF metric is $0.042$, which indicates our policy is close to being perfectly counterfactually fair. Indeed, the CF metric should be exactly 0 if we know the true dynamics of the environment of interest; the reason why it is not exactly 0 here is that we need to estimate the dynamics of the environment of interest during preprocessing, which can introduce finite-sample errors.

\subsection{Comparisons against Baseline Methods}

We can compare the sequential data preprocessing method in \texttt{PyCFRL} against a few baselines: "Random", which selects each action randomly with equal probability; "Full", which uses all variables, including the sensitive attribute, for policy learning; and "Unaware", which uses all variables except the sensitive attribute for policy learning. We implemented these baselines and evaluated their values and counterfactual unfairness levels as part of the code example of the "Assessing Policies Using Real Data" workflow in the \href{https://pycfrl-documentation.netlify.app/tutorials/example_workflows}{"Example Workflows"} section of the \texttt{PyCFRL} documentation. We summarize below the values and CF metrics calculated in this code example, where "Ours" stands for outputs from the \texttt{SequentialPreprocessor}.

\begin{table}[H]
\centering
\begin{tabularx}{\textwidth}{lYYYY}
\toprule
 & Random & Full & Unaware & Ours \\
\midrule
Value & $-1.444$ & $8.606$ & $8.588$ & $7.358$ \\
Counterfactual Unfairness Level & $0$ & $0.407$ & $0.446$ & $0.042$ \\
\bottomrule
\end{tabularx}
\caption{Comparisons of trained policies against baseline methods.}
\label{tab:baseline_comparison}
\end{table}

By definition, the "Random" baseline always achieves perfect CF. On the other hand, "Ours" resulted in much fairer policies than "Full" and "Unaware", which suggests that the \texttt{SequentialPreprocessor} can effectively control counterfactual unfairness. Nevertheless, as a trade-off for better CF, "Ours" achieved a lower value than "Full" and "Unaware".

\section{Conclusions}

\texttt{PyCFRL} is a Python library that enables counterfactually fair reinforcement
learning through data preprocessing. It also provides tools to calculate
the value and unfairness level of a given policy. To our knowledge, it is the first library to address CF
problems in the context of RL. The practical utility of \texttt{PyCFRL} can be further improved via extensions. First, the current \texttt{PyCFRL} implementation requires every individual in the offline dataset to
have the same number of time steps. Extending the library to accommodate
variable-length episodes can improve its flexibility and usefulness. Second, \texttt{PyCFRL} can further combine the preprocessor with popular offline RL algorithm libraries such as
\texttt{d3rlpy}~\citep{d3rlpy}, or connect the evaluation functions with established RL
environment libraries such as \texttt{gym}~\citep{towers2024gymnasium}. Third, generalization to non-additive counterfactual states reconstruction can make \texttt{PyCFRL} more versatile. We leave these extensions 
to future updates.

\section*{Acknowledgements}

Jianhan Zhang and Jitao Wang contributed equally to this work. The authors declare no conflicts of interest.

\bibliographystyle{apalike}
\bibliography{references}

\end{document}